\documentclass[review]{elsarticle}

\usepackage{lineno,hyperref}
\usepackage{longtable}
\modulolinenumbers[5]

\newcommand\cites[1]{\citeauthor{#1}'s\ (\citeyear{#1})}

\biboptions{authoryear}

\begin{document}

\begin{frontmatter}

\title{First and Second Order Dynamics in a Hierarchical SOM system for Action Recognition}

\author{Zahra Gharaee, Peter G\"ardenfors and Magnus Johnsson}
\address{Lund University Cognitive Science,\\
Helgonav\"agen 3, 221 00 Lund, Sweden}
\ead{zahra.gharaee@lucs.lu.se, peter.gardenfors@lucs.lu.se,magnus@magnusjohnsson.se}

\begin{abstract}
Human recognition of the actions of other humans is very efficient and is based on patterns of movements. Our theoretical starting point is that the dynamics of the joint movements is important to action categorization. On the basis of this theory, we present a novel action recognition system that employs a hierarchy of Self-Organizing Maps together with a custom supervised neural network that learns to categorize actions. The system preprocesses the input from a Kinect like 3D camera to exploit the information not only about joint positions, but also their first and second order dynamics. We evaluate our system in two experiments with publicly available data sets, and compare its performance to the performance with less sophisticated preprocessing of the input. The results show that including the dynamics of the actions improves the performance. We also apply an attention mechanism that focuses on the parts of the body that are the most involved in performing the actions. 
\end{abstract}


\end{frontmatter}

\section{Introduction}
The success of human-robot interaction depends on the development of robust methods that enable robots to recognize and predict goals and intentions of other agents. Humans do this, to a large extent, by interpreting and categorizing the actions they perceive. Hence, it is central to develop methods for action categorization that can be employed in robotic systems. This involves an analysis of on-going events from visual data captured by cameras to track movements of humans and to use this analysis to identify actions. One crucial question is to know what kind of information should be extracted from observations for an artificial action recognition system.

Our ambition is to develop an action categorization method that, at large, works like the human system. We present a theory of action categorization due to \citet{Gardenfors2} (see also \citet{Gardenfors1} and \citet {Gardenfors5}) that builds on \cites{Gardenfors4} theory of conceptual spaces. The central idea is that actions are represented by the underlying force patterns. Such patterns can be derived from the second order dynamics of the input data. We present experimental data on how humans categorize action that supports the model. A goal of this article is to show that if the dynamics of actions is considered, the performance of our Self-Organizing Maps (SOMs) \citep{kohonen} based action recognition system can be improved when categorizing actions based on 3D camera input.

The architecture of our action recognition system is composed of a hierarchy of three neural network layers. These layers have been implemented in different versions. The first layer consists of a SOM, which is used to represent preprocessed input frames (e.g. posture frames) from input sequences and to extract their motion patterns. This means that the SOM reduces the data dimensionality of the input and the actions in this layer are represented as activity patterns over time.

The second layer of the architecture consists of a second SOM. It receives the superimposed activities in the first layer for complete actions. The superimposition of all the activity in the first layer SOM provides a mechanism that makes the system time invariant. This is because similar movements carried out at different speed elicit similar sequences of activity in the first layer SOM. Thus the second layer SOM represents and clusters complete actions. The third layer consists of a custom made supervised neural network that labels the different clusters in the second layer SOM with the corresponding action.

We have previously studied the ability of SOMs to learn discriminable representations of actions \citep{Buonamente4}, and we have developed a hierarchical SOM based action recognition architecture. This architecture has previously been tested using video input from human actions in a study that also included a behavioural comparison between the architecture and humans \citep{Buonamente3}, and using extracted joint positions from a Kinect like 3D camera as input with good results \citep{Gharaee3}. 

This article presents results that suggest that the performance of our action recognition architecture can be improved by exploiting not only the joint positions extracted from a Kinect like 3D camera, but also simultaneously the information present in their first and second order dynamics.

Apart from analysing the dynamics of the data, we implement an attention mechanism that is inspired by how human attention works. We model attention by reducing the input data to those parts of the body that contribute the most in performing the various actions. Adding such an attention mechanism improves the performance of the system. 

The rest of the paper is organized as follows: First we present the theoretical background from cognitive science in section 2. The action recognition architecture is described in detail in section 3. Section 4 presents two experiments to evaluate the performance of the architecture employing new kinds of preprocessing to enable additional dynamic information as additional input. Section 5 concludes the paper.

\section{Theoretical Background}
When investigating action recognition in the context of human-robot interaction, it should first be mentioned that human languages contain two types of verbs describing actions \citep{Levin, Warglien}. The first type is manner verbs that describe how an action is performed. In English, some examples are ‘run’, ‘swipe’, ‘wave’, ‘push’, and ‘punch’. The second type is result verbs that describe the result of actions. In English, some examples are ‘move’, ‘heat’, ‘clean’, ‘enter’, and ‘reach’.

In the context of robotics, research has focused on how result verbs can be modelled (e.g. \citet{Cangelosi}, \citet{Kalkan}, \citet{Lallee} and \citet{Demiris}). However, when it comes to human-robot interaction, the robot should also be able to recognize human actions by the manner they are performed. This is often called recognition of biological motion \citep{Hemeren}. Recognising manner action is important in particular if the robot is supposed to model the intentions of a human. In the literature, there are some systems for categorizing human actions described by manner verbs, e.g. \citet{Giese2} and \citet{Giese3}. However, these systems have not been developed with the aim of supporting human-robot interaction. Our aim in this article is to present a system that recognises a set of manner actions. Our future aim is, however, to integrate this with a system for recognising results verbs that can be used in linguistic interactions between a human and an robot (see \citet{Cangelosi}, \citet{Mealier} for examples of such linguistic systems).

Results from the cognitive sciences indicate that the human brain performs a substantial information reduction when categorizing human manner actions. \citet{johansson} has shown that the kinematics of a movement contain sufficient information to identify the underlying dynamic patterns. He attached light bulbs to the joints of actors who were dressed in black and moved in a black room. The actors were filmed performing actions such as walking, running, and dancing. Watching the films - in which only the dots of light could be seen - subjects recognized the actions within tenths of a second. Further experiments by \citet{Runesson1}, see also \citep{Runesson2}, show that subjects extract subtle details of the actions performed, such as the gender of the person walking or the weight of objects lifted (where the objects themselves cannot be seen).

One lesson to learn from the experiments by Johansson and his followers is that the kinematics of a movement contains sufficient information to identify the underlying dynamic force patterns. \citet{Runesson2} claims that people can directly perceive the forces that control different kinds of motion. He formulates the following thesis:

\textit{Kinematic specification of dynamics}: The kinematics of a movement contains sufficient information to identify the underlying dynamic force patterns.
 
From this perspective, the information that the senses - primarily vision - receive about the movements of an object or individual is sufficient for the brain to extract, with great precision, the underlying forces. Furthermore, the process is automatic: one cannot help but perceiving the forces.

Given these results from perceptual psychology, the central problem for human-robot interaction now becomes how to construct a model of action recognition that can be implemented in a robotic system. One idea for such a model comes from \citet {Marr2} and \citet {Vaina}, who extend \cites{Marr1} cylinder models of objects to an analysis of actions. In Marr's and Vaina's model, an action is described via differential equations for movements of the body parts of, for example, a walking human. What we find useful in this model is that a cylinder figure can be described as a vector with a limited number of dimensions. Each cylinder can be described by two dimensions: length and radius. Each joining point in the figure can be described by a small number of coordinates for point of contact and angle of joining cylinder. This means that, at a particular moment, the entire figure can be written as a (hierarchical) vector of a fairly small number of dimensions. An action then consists of a sequence of such vectors. In this way, the model involves a considerable reduction of dimensionality in comparison to the original visual data. Further reduction of dimensionality is achieved in a skeleton model.

It is clear that, using Newtonian mechanics, one can derive the differential equations from the forces applied to the legs, arms, and other moving parts of the body. For example, the pattern of forces involved in the movements of a person running is different from the pattern of forces of a person walking; likewise, the pattern of forces for saluting is different from the pattern of forces for throwing \citep{Vaina2}.

The human cognitive apparatus is not exactly evolved for Newtonian mechanics. Nevertheless, \citet{Gardenfors1} (see also \citet{Warglien} and \citet{Gardenfors5}) proposed that the brain extracts the forces that lie behind different kinds of movements and other actions:

\textit{Representation of actions}: An action is represented by the pattern of forces that generates it. 

We speak of a pattern of forces since, for bodily motions, several body parts are involved; and thus, several force vectors are interacting (by analogy with Marr's and Vaina's differential equations). Support for this hypothesis will be presented below. One can represent these patterns of forces in principally the same way as the patterns of shapes described in \citet{Gardenfors5}, section 6.3. In analogy with shapes, force patterns also have meronomic structure. For example, a dog with short legs moves in a different way than a dog with long legs.

This representation fits well into the general format of conceptual spaces presented by \citet {Gardenfors4, Gardenfors1}. In order to identify the structure of the action domain, similarities between actions should be investigated. This can be accomplished by basically the same methods used for investigating similarities between objects. Just as there, the dynamic properties of actions can be judged with respect to similarities: for example, walking is more similar to running than to waving. Very little is known about the geometric structure of the action domain, except for a few recent studies that we will present below. We assume that the notion of betweenness is meaningful in the action domain, allowing us to formulate the following thesis in analogy to the thesis about properties (see \citet{Gardenfors4, Gardenfors1} and \citet{Gardenfors2}):

\textit{Thesis about action concepts}: An action concept is represented as a convex region in the action domain.

One may interpret here convexity as the assumption that, given two actions in the region of an action concept, any linear morph between those actions will fall under the same concept. 

One way to support the analogy between how objects and how actions are represented in conceptual space is to establish that action concepts share a similar structure with object categories \citep[p. 25]{Hemeren}. Indeed, there are strong reasons to believe that actions exhibit many of the prototype effects that \citet {Rosch} presented for object categories. In a series of psychological experiments, \citet {Hemeren} showed that action categories show a similar hierarchical structure and have similar typicality effects to object concepts. He demonstrated a strong inverse correlation between judgements of most typical actions and reaction time in a word/action verification task.

Empirical support for the thesis about action concepts as regards body movements can also be found in \citet {Giese2}. Using \cites{johansson} patch-light technique, they started from video recordings of natural actions such as walking, running, limping, and marching. By creating linear combinations of the dot positions in the videos, they then made films that were morphs of the recorded actions. Subjects watched the morphed videos and were asked to classify them as instances of walking, running, limping, or marching, as well as to judge the naturalness of the actions. Giese and Lappe did not explicitly address the question of whether the actions recognized form convex regions in the force domain. However, their data clearly support this thesis.

Another example of data that can be used to study force patterns comes from \citet{Wang}. They collected data from the walking patterns of humans under different conditions. Using the methods of \citet{Giese3}, these patterns can be used to calculate the similarity of the different gaits.

A third example is \citet{Malt} who studied how subjects named the actions shown in 36 video clips of different types of walking, running and jumping. The subjects were native speakers of English, Spanish, Dutch, and Japanese. The most commonly produced verb for each clip in each language was calculated. This generated a number of verbs also including several subcategories. Another group of subjects, again native speakers of the four languages, judged the physical similarity of the actions in the video clips. Based on these judgements a two-dimensional multidimensional scaling solution was calculated. The verbs from the first group were then mapped onto this solution. Figure 4 in \citet{Malt} shows the results for the most common English action word for each video clip. The results support the thesis that regions corresponding to the names are convex.

The action recognition system presented in this article has similarities with these models in the sense that actions are represented as sequences of vectors, and it categorizes the actions on basis of their similarities. In our system, similarity is modelled as closeness in SOMs. We next turn to a description of the architecture of the system.

\section{Hierarchical SOM Architecture for Action Recognition}
\begin{figure}%
\begin{center}
\begin{minipage}{120mm}
\centering%
\includegraphics[width=4.0in]{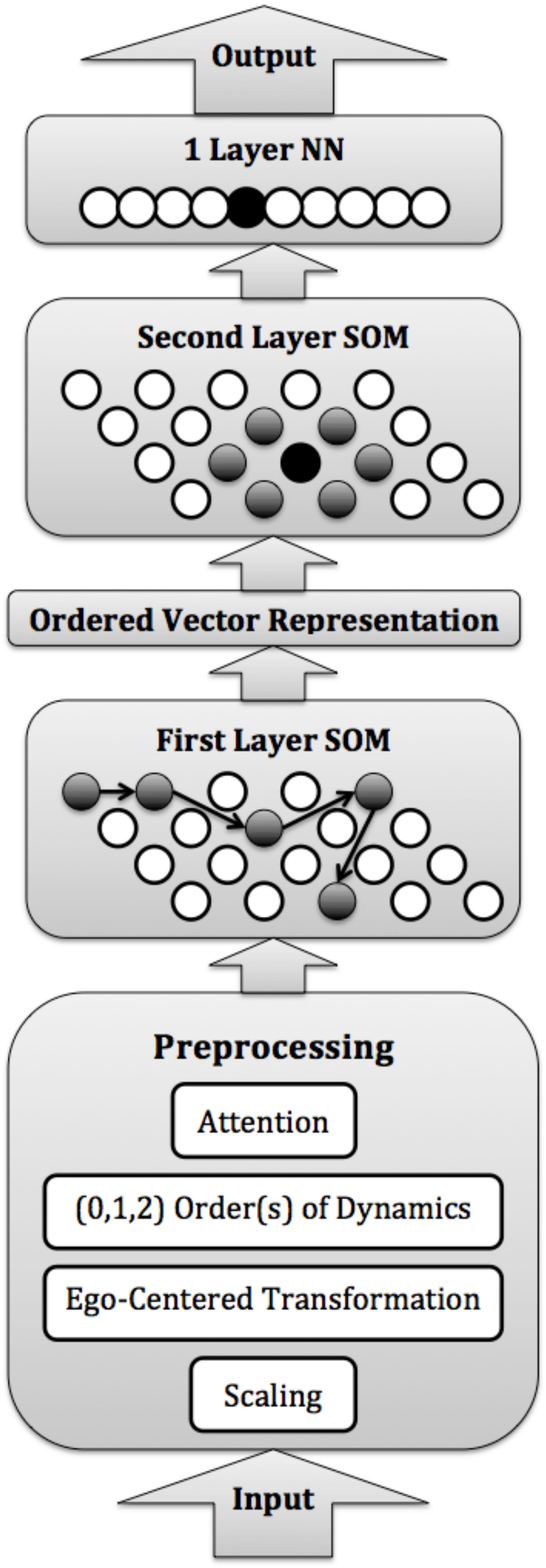}\hspace{5pt}
\caption
{The architecture of the action recognition system, consisting of three layers where the first and second layers are SOMs, and the third layer is a custom supervised neural network. The input is a combination of joint positions extracted from the output of a Kinect like depth camera.}
\label{fig:SOM_SOM_OutputLayer}
\end{minipage}
\end{center}
\end{figure}

Our action recognition system consists of a three layered neural network architecture (Fig.~\ref{fig:SOM_SOM_OutputLayer}). The first layer consists of a SOM that develops an ordered representation of preprocessed input. The second layer consists of a second SOM that receives, as input, the superimposed sequence of activity elicited during an action in the first layer SOM. Thus the second layer SOM develops an ordered representation of the activity traces in the first layer SOM that correspond to different actions. Finally, the third layer is a custom supervised neural network that associates activity representing action labels to the activity in the second layer SOM.

\subsection{Preprocessing}
From a computational point of view, there are many challenges that make the action recognition task difficult to imitate artificially. For example, the acting individuals differ in height, weight and bodily proportions. Other important issues to be addressed are the impact of the camera's viewing angle and distance from the actor and the performance speed of the actions. In brief, categorizations of actions ought to be invariant under distance, viewing angle, size of the actor, lighting conditions and temporal variations.

\subsubsection{Ego-Centered Transformation}
\begin{figure}%
\begin{center}
\begin{minipage}{120mm}
\centering%
\includegraphics[width=4.00in]{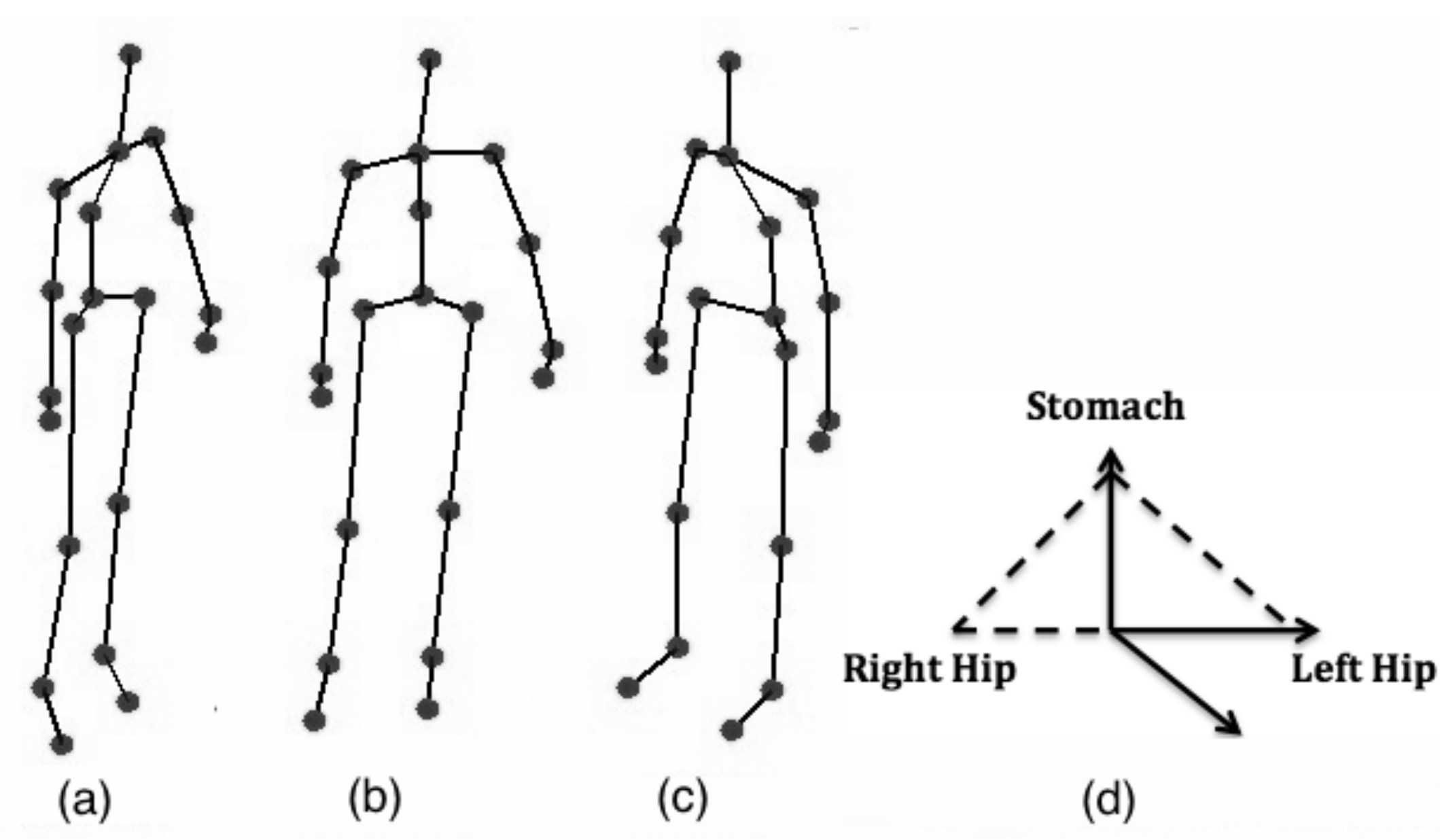}\hspace{5pt}
\caption
{Different body orientations; turned to the left (a), front direction (b), turned to the right (c) and the joints used to calculate the egocentric coordinate system (d). }
\label{fig:ego-cent2}
\end{minipage}
\end{center}
\end{figure}

To make the action recognition system invariant to different orientations of the action performing agents, coordinate transformation into an ego-centered coordinate system, Fig.~\ref{fig:ego-cent2}, located in the central joint, stomach in Fig.~\ref{fig:Skeleton}, is applied to the extracted joint positions. To build this new coordinate system three joints named Right Hip, Left Hip and Stomach are used.

\subsubsection{Scaling}
\begin{figure}%
\begin{center}
\begin{minipage}{120mm}
\centering%
\includegraphics[width=2.00in]{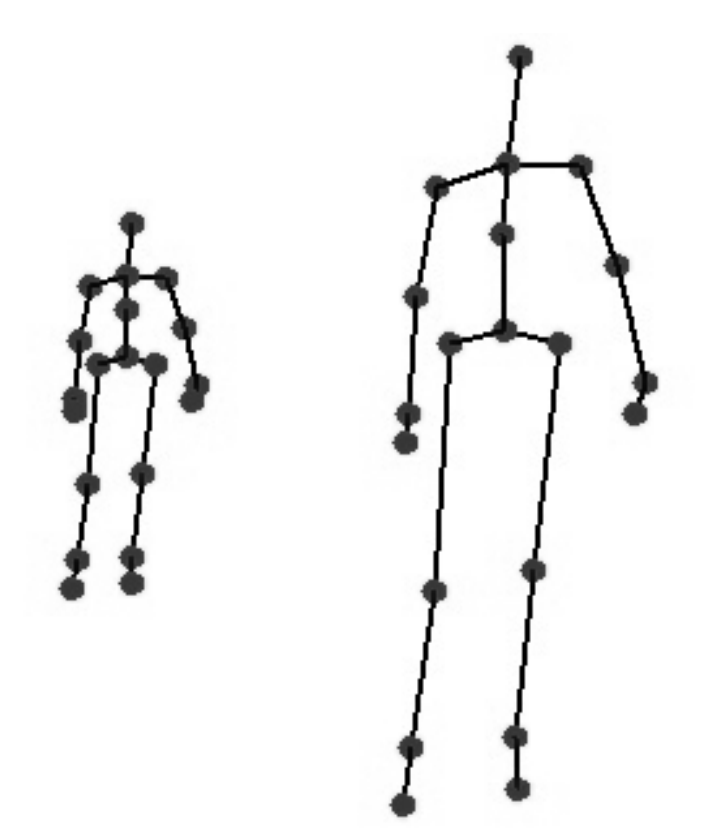}\hspace{5pt}
\caption
{Different skeleton sizes due to different distances from the depth camera.}
\label{fig:re-scale}
\end{minipage}
\end{center}
\end{figure}

The other preprocessing mechanism applied to the joint positions is the scaling transformation to make the system invariant to the distance from the depth camera. Due to the scaling to a standard size, the representations of the agent will always have a fixed size even if the action performers have different distances to the depth camera, Fig.~\ref{fig:re-scale}.

\subsubsection{Attention}
As humans we learn how to control our attention in performing other tasks \cite{Shariatpanahi}. Utilizing an attention mechanism can, together with other factors (e.g, state  estimation), improve  our performance in doing many other tasks \cite{Gharaee1}. One of the strongest factors that attract our attention is the movement. Here, we therefore assume that when actions are performed, the observer pays attention to the body parts that are most involved in the actions. Thus, in the experiments we have applied attention mechanisms, that direct the attention towards the body parts that move the most. Thus, for example, when the agent is clapping hands, the attention is focused on the arms and no or very little attention is directed towards the rest of the agent's body. 

Our attention mechanisms are simulated by only extracting some of the joint coordinates. These coordinates are determined by how much they are involved in performing the actions. These mechanisms reduce the number of input dimensions that enters the first-layer SOM, and it helps the system to recognize actions by focusing on the more relevant parts of the body while ignoring the less relevant parts. This procedure increases the accuracy of the system. The attention mechanism applied to the system will be described in the experiment section below. 

\subsubsection{Dynamics}
By using postures composed of 3D joints positions as input to our architecture, a high performance of our action recognition system has been obtained \citep{Gharaee3}. In addition to the 3D positioning, our body joints have velocity and acceleration that could be modeled as the first and second orders of dynamics. The first order of dynamic (velocity) determines the speed and the direction of joint's movements while the acceleration, via Newton’s second law, determines the direction of force vector applies to joint during acting.

In this study, the first and second order joints dynamics have been extracted and used together with the 3D positions. This has been done by calculating the differences between consecutive sets of joint positions, and between consecutive sets of first order joint dynamics in turn. This has enabled us to investigate how the inclusion of the dynamics contributes to the performance of our action recognition system.

\subsection{The First and Second Layer SOMs}
The first two layers of our architecture consist SOMs. The SOMs self-organizes into dimensionality reduced and discretized topology preserving representations of their respective input spaces. Due to the topology preserving property nearby parts of the trained SOM respond to similar input patterns. This is reminiscent of the cortical maps found in mammalian brains. The topology-preserving property of SOMs is a consequence of the use of a neighbourhood function during the adaptation of the neuron responses, which means the adaptation strength of a neuron is a decreasing function of the distance to the most activated neuron in the SOM. This also provides the SOM, and in the extension our action recognition system, with the ability to generalize learning to novel inputs, because similar inputs elicit similar activity in the SOM. Thus similar actions composed of similar sequences of agent postures and dynamics will elicit similar activity trajectories in the first layer SOM, and similar activity trajectories in the first layer SOM, and thus similar actions, will be represented nearby in the second layer SOM. Since a movement performed at different speeds will elicit activity along the same trajectory in the first layer SOM when the input consists of a stream of sets of joint positions, as well as when dynamics is added (as long as the action's internal dynamic relations are preserved), our action recognition system also achieves time invariance.  

The SOM consists of an $I\times J$ grid of neurons with a fixed number of neurons and a fixed topology. Each neuron $n_{ij}$ is associated with a weight vector $w_{ij}\in{R}^n$ with the same dimensionality as the input vectors. All the elements of the weight vectors are initialized by real numbers randomly selected from a uniform distribution between 0 and 1, after which all the weight vectors are normalized, i.e. turned into unit vectors.

At time $t$ each neuron $n_{ij}$ receives the input vector $x(t)\in{R}^n$.
The net input $s_{ij}(t)$ at time $t$ is calculated using the Euclidean metric:

\begin{equation}
s_{ij}(t)=||x(t) - w_{ij}(t)||
\end{equation}

The activity $y_{ij}(t)$ at time $t$ is calculated by using the exponential function:

\begin{equation}
y_{ij}(t)=e^{\frac{-s_{ij}(t)}{\sigma}}
\end{equation}

The parameter $\sigma$ is the exponential factor set to ${10^6}$  and $0 \leq {i} < I$, $0 \leq {j} < J$. ${i},{j}\in{N}$. The role of the exponential function is to normalize and increase the contrast between highly activated and less activated areas.

The neuron $c$ with the strongest activation is selected:

\begin{equation}
c=\mathrm {arg} \mathrm{ max}_{ij}y_{ij}(t)
\end{equation}

The weights $w_{ijk}$ are adapted by

\begin{equation}
w_{ijk}(t+1)=w_{ijk}(t)+\alpha(t)G_{ijc}(t)[x_k(t)-w_{ijk}(t)]
\end{equation}

The term $0 \leq \alpha(t) \leq 1$ is the adaptation strength, $\alpha(t) \rightarrow 0$ when $t \rightarrow \infty$. The neighbourhood function $G_{ijc}(t) = e^{-\frac{||r_c - r_{ij}||}{2\sigma^2(t)}}$ is a Gaussian function decreasing with time, and $r_c \in R^2$ and $r_{ij} \in R^2$ are location vectors of neurons $c$ and $n_{ij}$ respectively. All weight vectors $w_{ij}(t)$ are normalized after each adaptation.

\begin{figure}%
\begin{center}
\begin{minipage}{120mm}
\centering%
\includegraphics[width=3.50in]{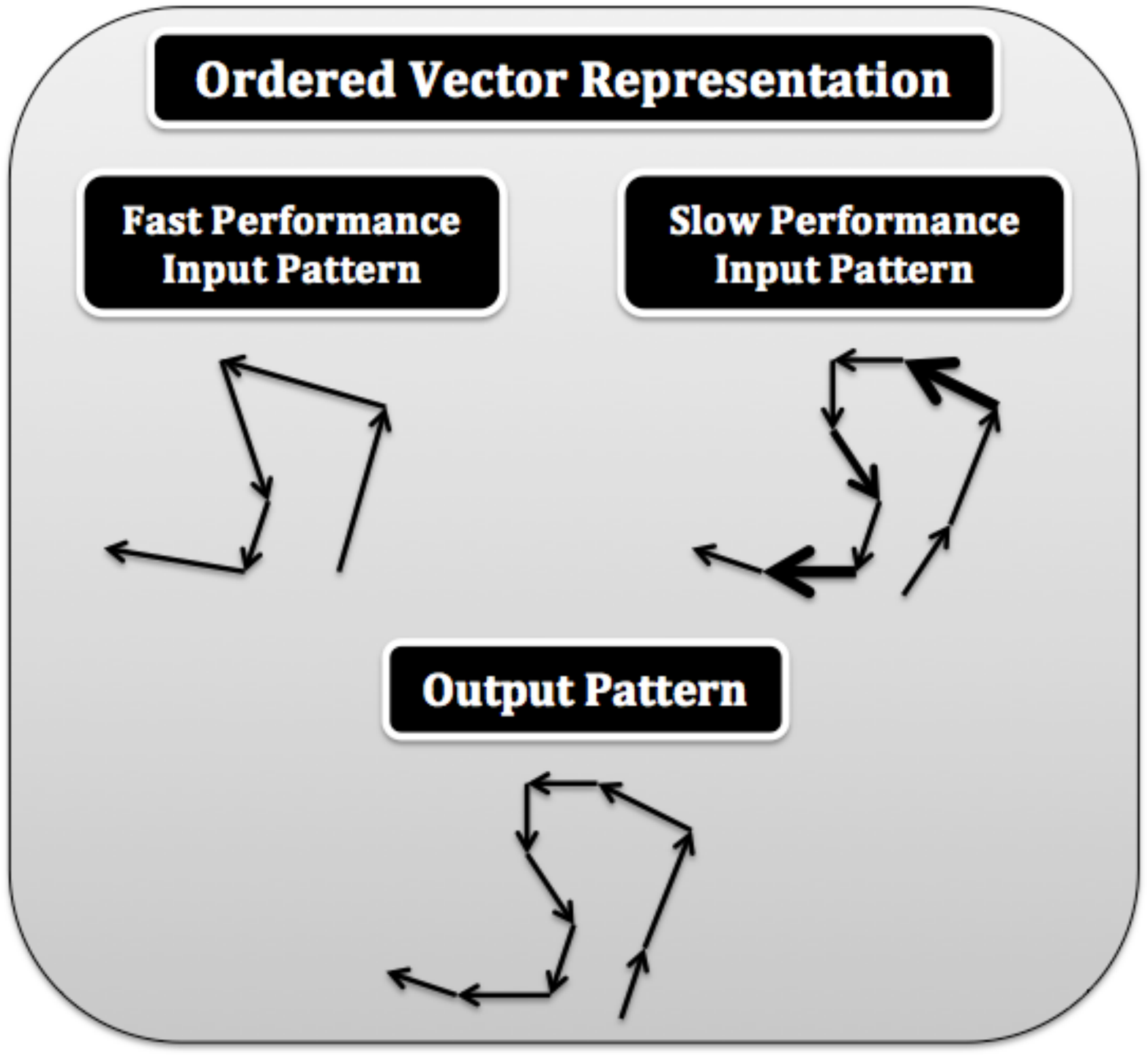}\hspace{5pt}
\caption
{Ordered Vector Representation Process. The figure shows the patterns in the first layer SOM elicited by the same action performed at different rates. When the action is performed slowly (right) there are more activations, a higher number of activated neurons and the same neurons may be activated repeatedly (the darker and larger arrows), than when the action is performed quickly (left). In both cases the activations are on the same path in the first layer SOM. The ordered vector representation process creates a representation of the path (lower) designed to be independent of the performance rate, thus achieving time invariance for the system.}
\label{fig:SupImp}
\end{minipage}
\end{center}
\end{figure}

\subsection{Ordered Vector Representation}
Ordered vector representation of the activity traces unfolding in the first-layer SOM during an action is a way to transform an activity pattern over time into a spatial representation which will then be represented in the second-layer SOM. In addition, as mentioned shortly above, it provides the system with time invariance. This means that if an action is carried out several times, but at different rates, the activity trace will progress along the same path in the first-layer SOM. If this path is coded in a consistent way, the system achieves a time invariant encoding of actions. If instead the activity patterns in the first-layer SOM were transformed into spatial representations by, for example, using the sequence of centres of activity in the first-layer SOM, time variance would not be achieved. A very quick performance of an action would then yield a small number of centres of activity along the path in the first-layer SOM, whereas a very slow performance would yield a larger number of centres of activity, some of them sequentially repeated in the same neurons (see Fig.~\ref{fig:SupImp}).

Strictly speaking, time invariance is only achieved when using sets of joint positions as input to the system, because then the first-layer SOM develops a topology preserving representation of postures. When using first or second order dynamics this is no longer the case, unless the action's internal dynamic relations are preserved, because then an action's representation in the first-order SOM will vary depending on the dynamics. On the other hand some discriminable information for the system is added. For discrimination performance, the benefits of adding the dynamics to the input seems to be to some extent bigger than the drawbacks.

The ordered vector representation process used in this study works as follows. The length of the activity trace of an action $\Delta_{j}$ is calculated by:

\begin{equation}
  \Delta_{j} = \sum_{i=1}^{N-1} {{||P_{i+1}-P_{i}||}_2}
\end{equation}

The parameter $N$ is the total number of centres of activity for the action sequence $j$ and $P_{i}$ is the $i$th centre of activity in the same action sequence. 

Suitable lengths of segments to divide the activity trace for action sequence $j$ in the first-layer SOM are calculated by:

\begin{equation}
  d_{j} = \Delta_{j}/N_{Max}
\end{equation}

The parameter $N_{Max}$ is the longest path in the first-layer SOM elicited by the $M$ actions in the training set. Each activity trace in the first-layer SOM, generated by an action, is divided into $d_{j}$ segments, and the coordinates of the borders of these segments in the order they appear from the start to the end on the activity trace are composed into a vector used as input to the second-layer SOM.

\subsection{Output Layer}
The output layer consists of an $I\times J$  grid of a fixed number of neurons and a fixed topology. Each neuron $n_{ij}$ is associated with a weight vector  $w_{ij}\in{R}^n$. All the elements of the weight vector are initialized by real numbers randomly selected from a uniform distribution between 0 and 1, after which the weight vector is normalized, i.e. turned into unit vectors.

At time \textit{t} each neuron $n_{ij}$ receives an input vector $x(t)\in{R}^n$.

The activity $y_{ij}(t)$ at time $t$ in the neuron $n_{ij}$ is calculated using the standard cosine metric:

\begin{equation}
y_{ij}(t)=\frac{x(t)\cdot w_{ij}(t)}{||x(t)||||w_{ij}(t)||}
\end{equation}

During the learning phase the weights $w_{ijl}$ are adapted by

\begin{equation}
w_{ijl}(t+1)=w_{ijl}(t)+\beta [y_{ij}(t) - d_{ij}(t)]
\end{equation}

The parameter  $\beta$ is the adaptation strength and $d_{ij}(t)$ is the desired activity for the neuron $n_{ij}$ at time $t$.

\section{Experiments}
\begin{figure}%
\begin{center}
\begin{minipage}{120mm}
\centering%
\includegraphics[width=3.00in]{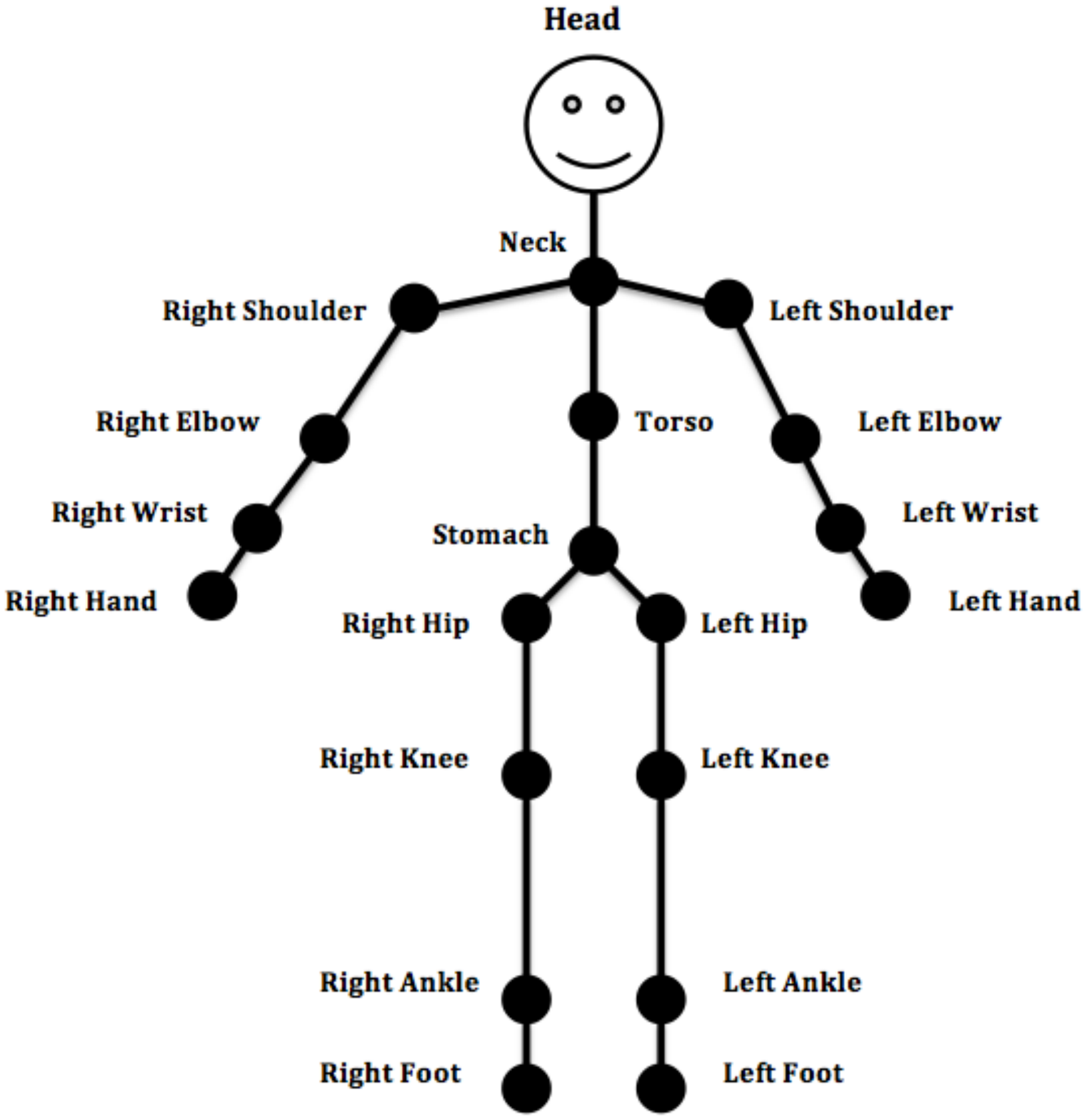}\hspace{5pt}
\caption
{The 3D information of the joints of the skeleton extracted from the 3D camera.}
\label{fig:Skeleton}
\end{minipage}
\end{center}
\end{figure}

We have evaluated the performance of our recognition architecture in two separate experiments using publicly available data from the repository MSR Action Recognition Datasets and Codes \citep {MSR}. Each experiment uses 10 different actions (in total 20 different actions) performed by 10 different subjects in 2 to 3 different events. The actions data are composed of sequences of sets of joint positions obtained by a depth camera, similar to a Kinect sensor. Each action sample is composed of a sequence of frames where each frame contains 20 joint positions expressed in 3D Cartesian coordinates as shown in Fig.~\ref{fig:Skeleton}. The sequences composing the action samples vary in length. We have used Ikaros framework \cite{balkenius} in order to design and implement our experiments.

\subsection{Experiment 1}
 In the first experiment, we have selected a set of actions from MSR Action3D Dataset containing 276 samples of 10 different actions performed by 10 different subjects, each in 2 to 3 different events. The actions of first experiment can be described as: 1. High Arm Wave, 2. Horizontal Arm Wave, 3. Using Hammer, 4. Hand Catch, 5. Forward Punch, 6. High Throw, 7. Draw X, 8. Draw Tick, 9. Draw Circle, 10. Tennis Swing. The first subset of actions was split into a training set containing 80\% of the action instances randomly selected from the original dataset and a test set containing the remaining 20\% of the action instances. 

\begin{figure}%
\begin{center}
\begin{minipage}{120mm}
\centering%
\includegraphics[width=2.00in]{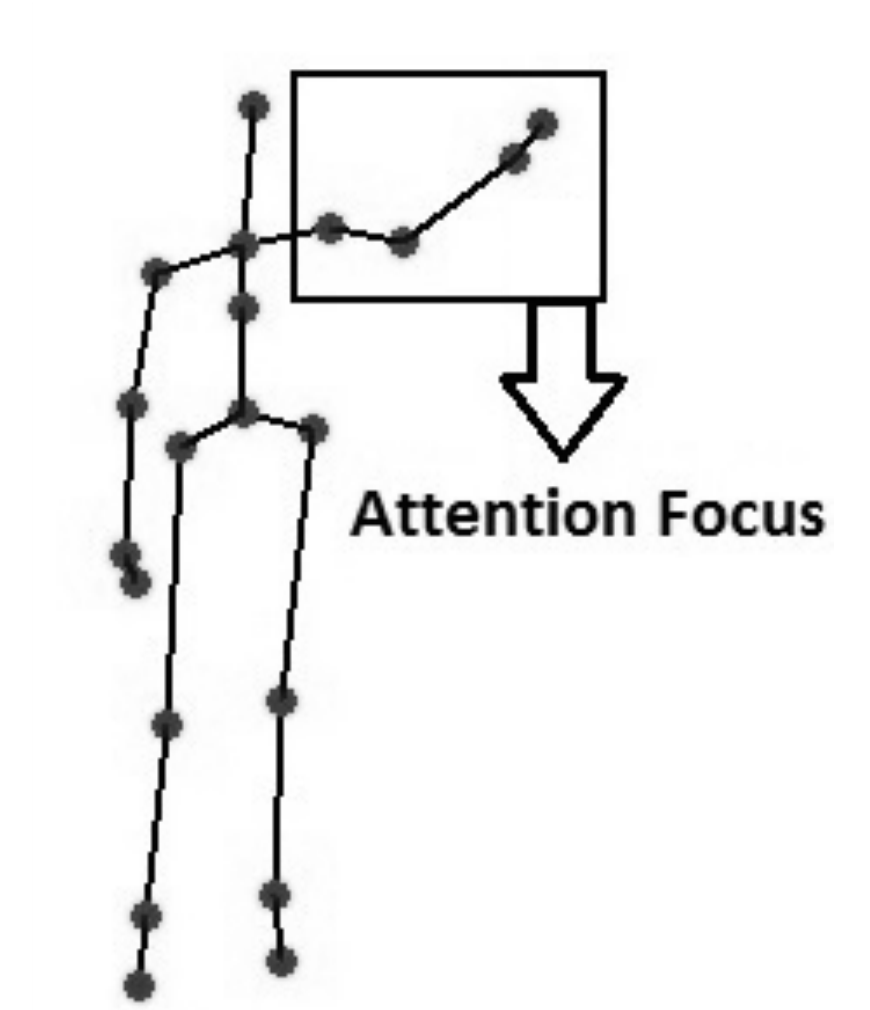}\hspace{5pt}
\caption
{The attention mechanism in the first experiment is obtained by setting the focus of attention to the left arm which is involved in performing all of the actions.}
\label{fig:Attention}
\end{minipage}
\end{center}
\end{figure}

The attention mechanism applied in this experiment is set to focus the attention to the arm which is the part of the body mainly involved in performing all of the actions, that is, the left arm for this experiment, see Fig.~\ref{fig:Attention}. The action recognition architecture was trained with randomly selected instances from the training set in two phases, the first training the first-layer $30\times30$ neurons SOM, and the second training the second-layer $35\times35$ neurons SOM together with the output-layer containing $10$ neurons. 

\begin{figure}%
\begin{center}
\begin{minipage}{120mm}
\centering%
\includegraphics[width=5.00in]{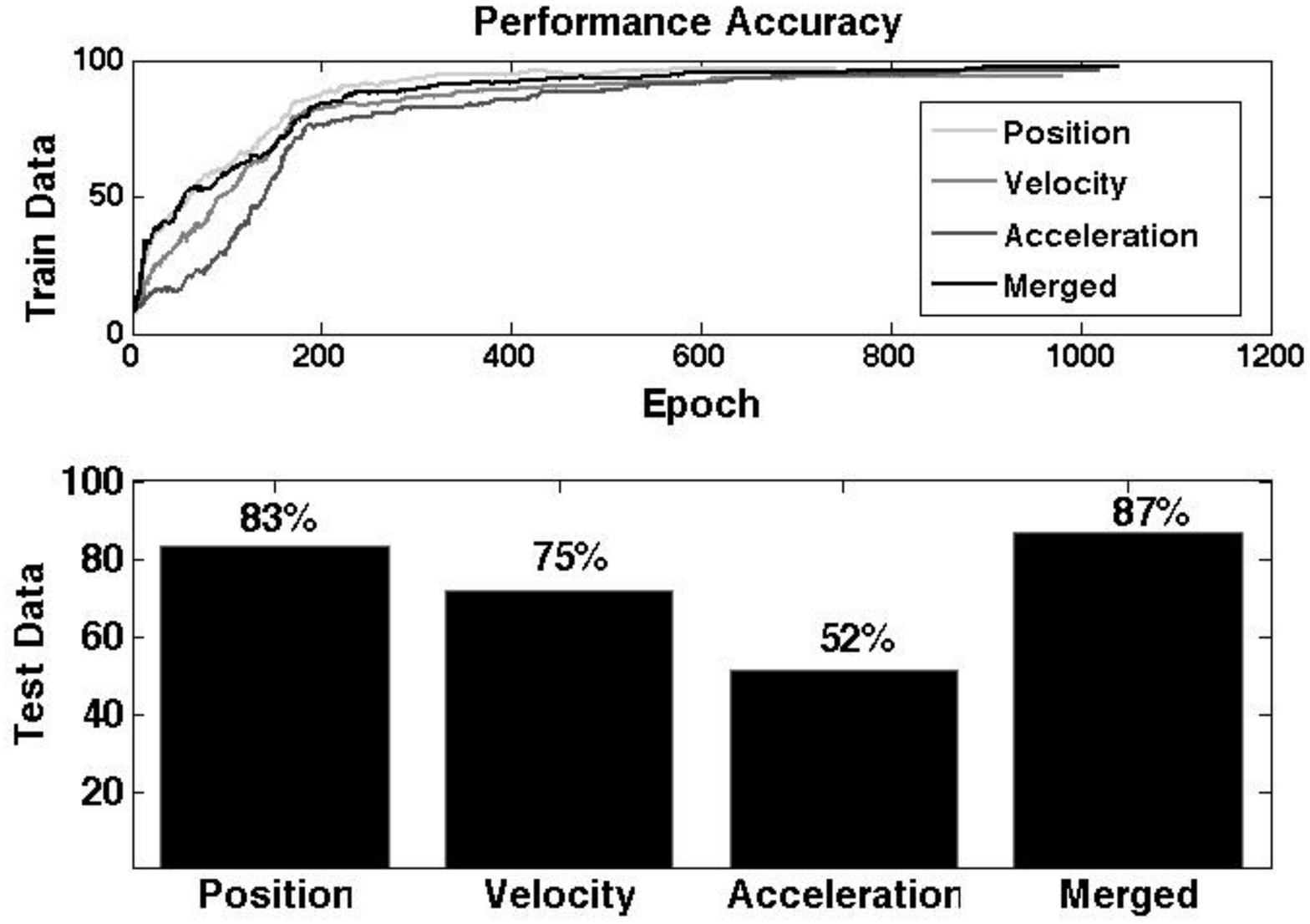}\hspace{5pt}
\caption
{Classification results of all actions when using as input only the joint positions, only the joint velocities (first derivatives), only the joint accelerations (second derivatives) or their combination (merged). Results with the training set during training (uppermost). Results for the fully trained system with the test set (lowermost). As can be seen, the best result was achieved when using the combined (merged) input.}
\label{fig:trtsall_}
\end{minipage}
\end{center}
\end{figure}

\begin{figure}%
\begin{center}
\begin{minipage}{120mm}
\centering%
\includegraphics[width=5.00in]{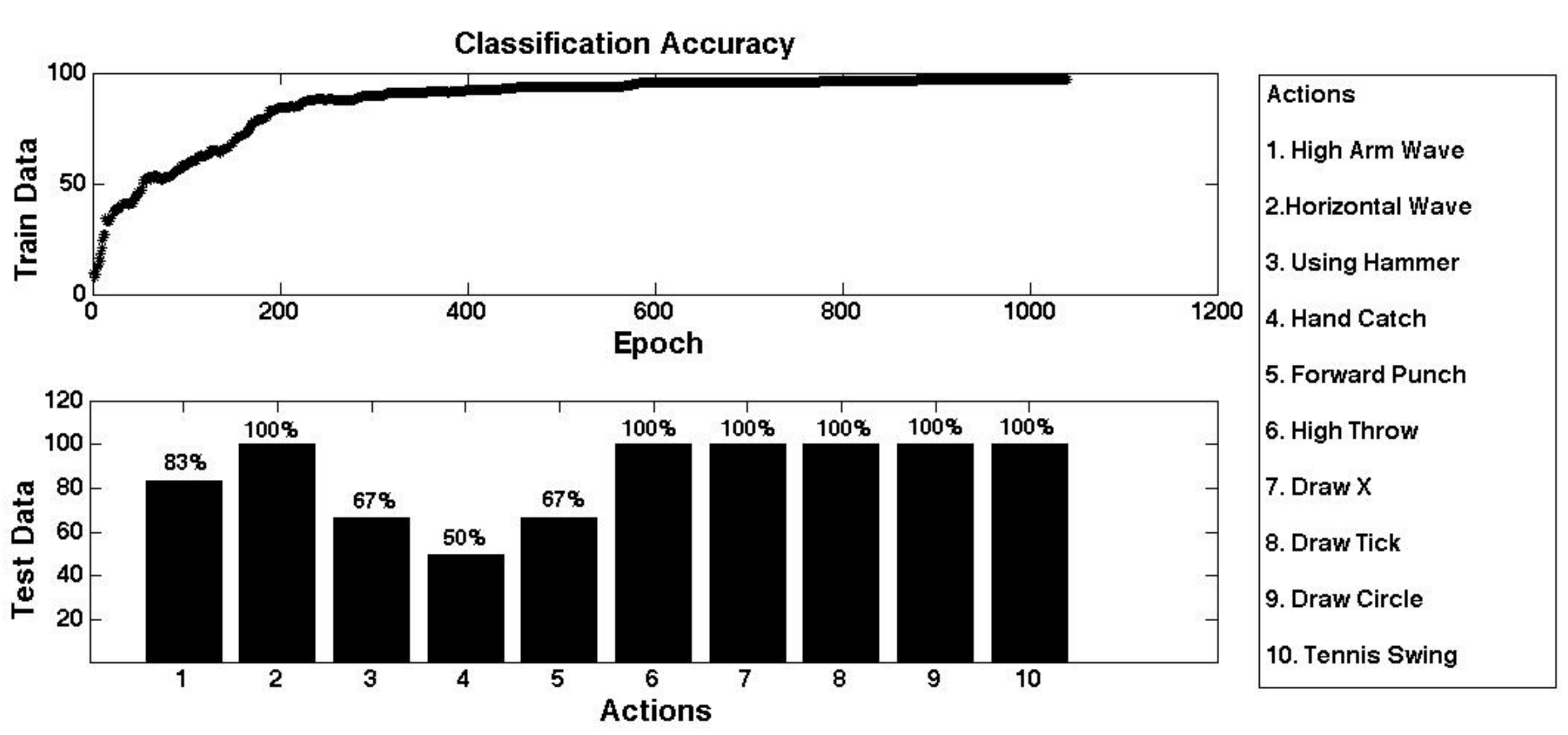}\hspace{5pt}
\caption
{Classification results of the action recognition system when receiving the combined input of joint positions and their first and second orders dynamics. Results with the training set during training (uppermost). Results for the fully trained system per action with the test set (lowermost).}
\label{fig:pvan_}
\end{minipage}
\end{center}
\end{figure}

In this experiment we continued the study of \cite{Gharaee3} in order to investigate the effect of applying dynamics on the system performance of action recognition. We achieved a recognition accuracy of 83\% for the postures (see  \cite{Gharaee3}). To extract the information of the first and second orders of dynamics we applied first the first order of dynamic (velocity) as the input to our system and obtained a recognition accuracy of 75\% and then the second order of dynamic (acceleration) as the input to our system where the recognition accuracy was 52\%. In the third step we merged the postures together with the first and second orders of dynamic (position, velocity and acceleration) as the input to the system (merged system). The categorization results show that 87\% of all test sequences are being correctly categorized. The results of using position, velocity or acceleration as input data one by one and then all together are depicted in Fig.~\ref{fig:trtsall_}. As the figure shows, when we used the combination of inputs in the merged system, 6 out of 10 actions were 100\% correctly categorized, see Fig.~\ref{fig:pvan_}.

\begin{figure}%
\begin{center}
\begin{minipage}{120mm}
\centering%
\includegraphics[width=5.00in]{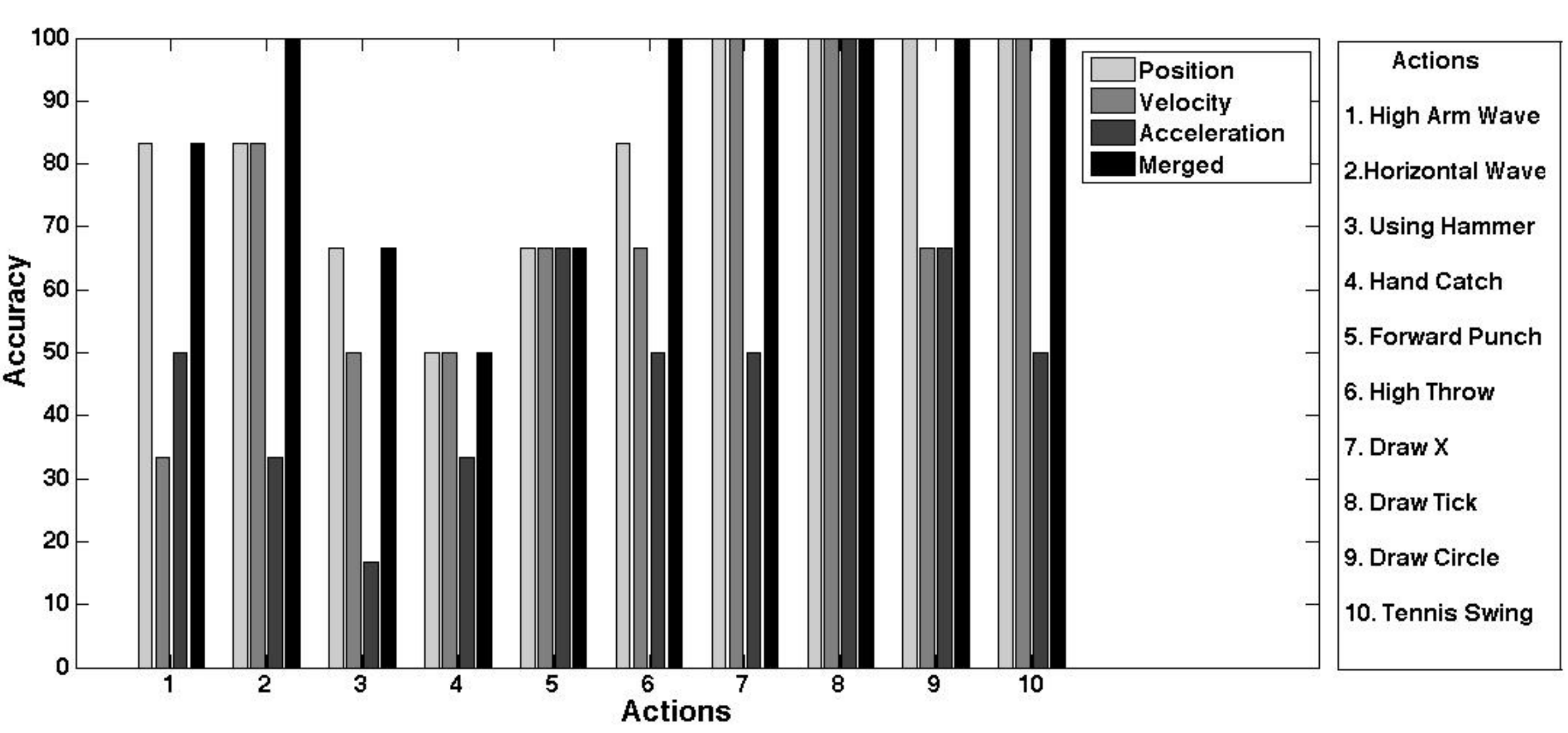}\hspace{5pt}
\caption
{Comparison of the classification performance, per action, of the action recognition system when using as input only the joint positions, only the joint velocities (first derivatives), only the joint accelerations (second derivatives) or their combination (merged).}
\label{fig:bars_}
\end{minipage}
\end{center}
\end{figure}

Fig.~\ref{fig:bars_} shows how the performance can be improved by using as input a combination of joint positions, joint velocities (first order of dynamic) and joint accelerations (second order of dynamic) compared to when each of these kinds of input is used alone.  

The results for the systems using only joint velocities and joints accelerations are not as good as the result for the system using only joint positions. One possible explanation for this is the low quality of the input data. The algorithm that extracts the skeleton data from the camera input is often not delivering biologically realistic results. The errors in joint positions that occur in the data set generated by the skeleton algorithm are magnified when the first derivatives are calculated for the joint velocities and doubly magnified when the second derivatives are calculated for the accelerations. We therefore believe that if our system was to be tested on a dataset with smaller errors, then the velocity and acceleration systems would perform better.

\subsection{Experiment 2}
In the second experiment we used the rest of actions in the MSR Action 3D Dataset. These actions are as follows: 1. Hand Clap, 2. Two Hand Wave, 3. Side Boxing, 4. Forward Bend, 5. Forward Kick, 6. Side Kick, 7. Jogging, 8. Tennis Serve, 9. Golf Swing, 10. Pick up and Throw. This second set of actions was split into a training set containing 75\% of the action instances randomly selected from the original dataset and a test set containing the remaining 25\% of the action instances. 

The action recognition architecture had the same settings for both experiments so the first layer SOM contained $30\times30$ neurons, and the second layer SOM contained $35\times35$ neurons and the output layer contained $10$ neurons. 

The attention mechanism used in this experiment is not as simple as in the first experiment because of the nature of actions of the second subset. These actions involve more varying parts of the body including the arms as well as the legs. Therefore extracting the body part that should form the focus of attention is not simple. Attention focus is determined, for each separate action, by a separate selection of the most moving body parts, which is inspired from human behaviour when observing the performing actions. 

For example, the action Forward Bend mainly involves the base part of the body which is composed of joints named  Head, Neck, Torso and Stomach, see Fig.~\ref{fig:Skeleton}, so the attention is focused on the base part, which includes the mentioned joints. Another example is the action Jogging which involves arms and legs so the attention is focused on the joints Left Ankle, Left Wrist, Right Ankle and Right wrist. For the actions named Hand Clap, Two Hand Wave and Side Boxing, Tennis Serve, Golf Swing and Pick up and Throw, the attention is focused on the joints Left Elbow, Left Wrist, Right Elbow and Right Wrist. Finally for the actions Forward Kick and Side Kick the attention is focused on the joints Left Knee, Left Ankle, Right Knee and Right Ankle. 

The attention mechanism significantly improves the performance of the system. Without using attention the system could only obtain an accuracy around 70\%. Therefore it is important to highlight the contribution of attention in developing the SOM architecture into a more accurate and optimal system.

\begin{figure}%
\begin{center}
\begin{minipage}{120mm}
\centering%
\includegraphics[width=5.00in]{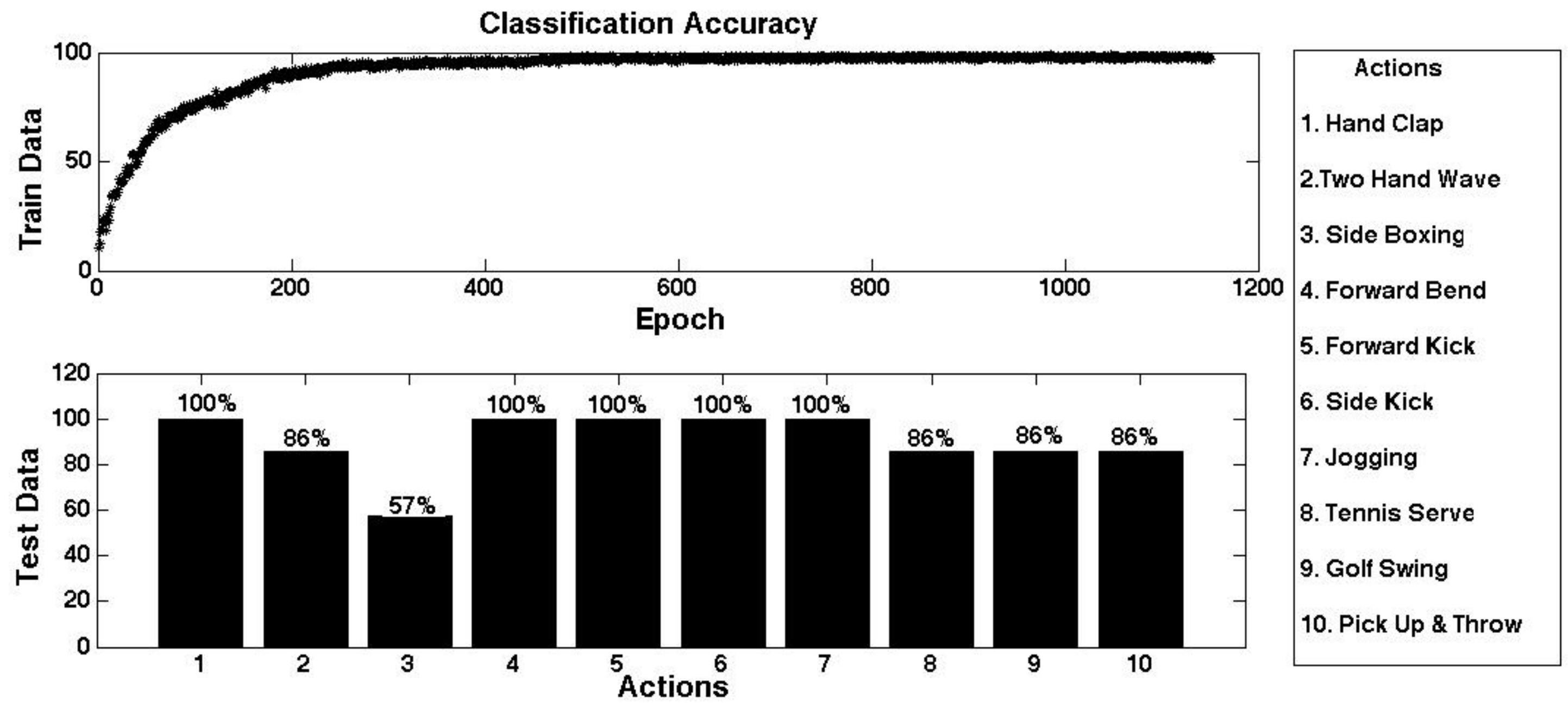}\hspace{5pt}
\caption
{Classification results of the action recognition system when receiving the combined input of joint positions and their first order dynamics (merged system). Results with the training set during training (uppermost). Results for the fully trained system, per action, with the test set (lowermost).}
\label{fig:pvan2nd}
\end{minipage}
\end{center}
\end{figure}

In the second experiment we first used only posture data as input to the architecture and reached a performance of 86\% correct recognitions of actions. In the next step we used merged input, i.e. posture data together with its first order dynamics (position and velocity). The categorization results show that 90\% of all test sequences was correctly categorized. As can be seen in Fig.~\ref{fig:pvan2nd}, 5 of the 10 actions are 100\% correctly categorized and the rest also have a very high accuracy. 

\begin{figure}%
\begin{center}
\begin{minipage}{120mm}
\centering%
\includegraphics[width=5.00in]{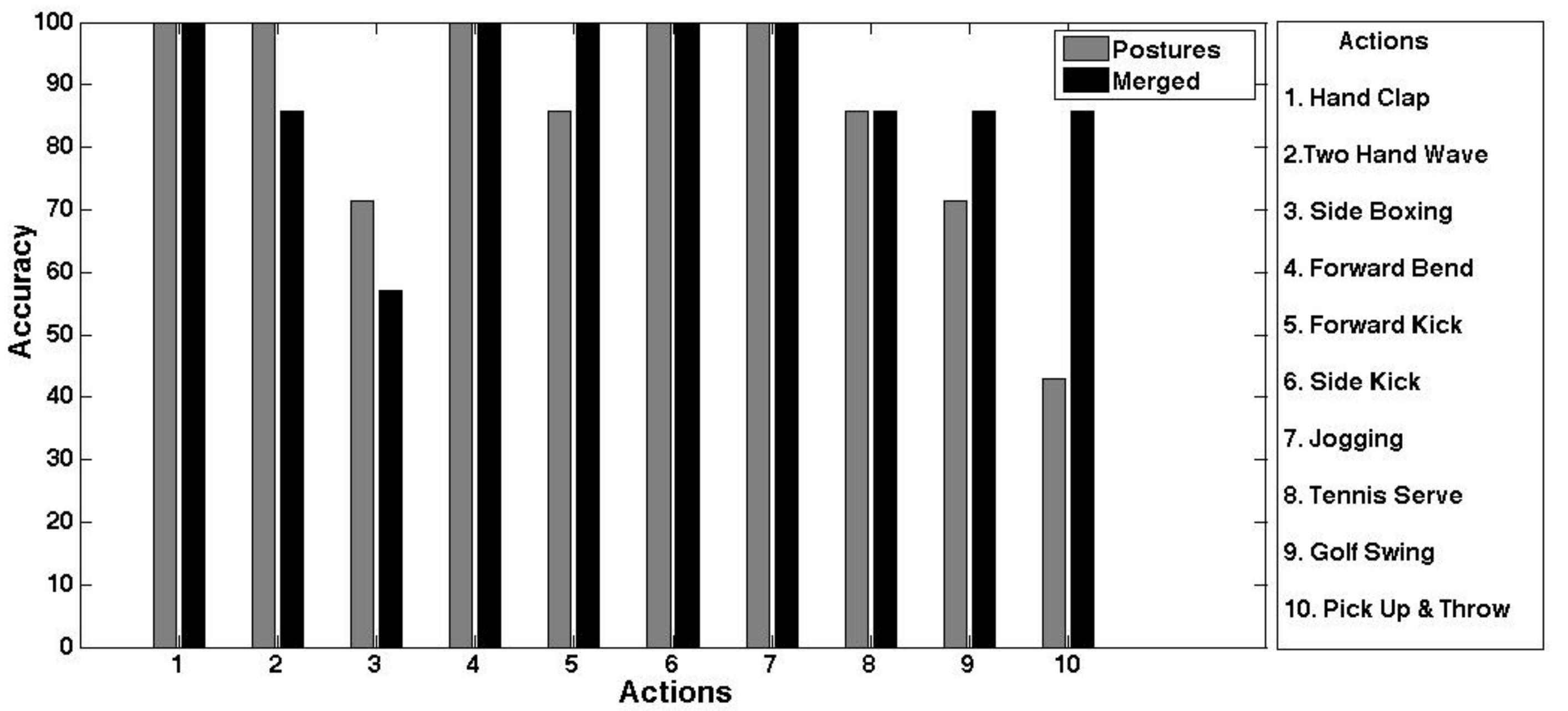}\hspace{5pt}
\caption
{Comparison of the classification performance, per action, of the action recognition system when using as input only the joint positions and the combination of the joint positions with  their first order dynamics (merged).}
\label{fig:bars2nd}
\end{minipage}
\end{center}
\end{figure}

Fig.~\ref{fig:bars2nd} shows a comparison of the performance when using only posture data (no dynamics) and when adding the dynamic to the system. Though the accuracy for the actions Two Hand Wave and Side Boxing is reduced when the dynamics is added to the system, the accuracy for the actions Forward Kick, Golf Swing and Pick Up and Throw are significantly improved. So in total the recognition performance was improved by adding the dynamics to the system.

\begin{figure}%
\begin{center}
\begin{minipage}{120mm}
\centering%
\includegraphics[width=4.70in]{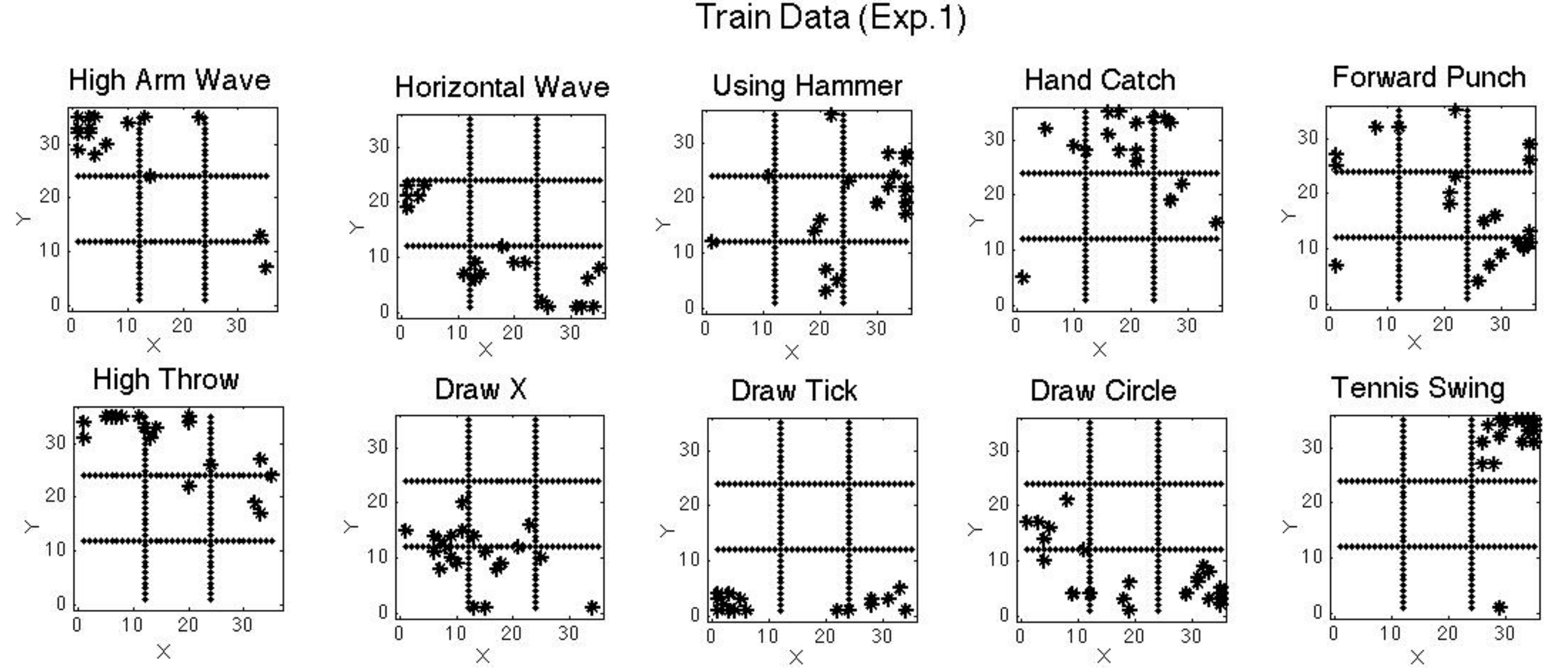}\hspace{5pt}
\caption
{Activations by the training set in the trained second layer SOM in the first experiment. The map is divided into 9 regions, starts from the bottom left square (region 1) and ends to the upper right square (region 9). As can be seen, the activations are spread out over several regions for many actions.}
\label{fig:trSOMBMap1}
\end{minipage}
\end{center}
\end{figure}

\begin{figure}%
\begin{center}
\begin{minipage}{120mm}
\centering%
\includegraphics[width=4.70in]{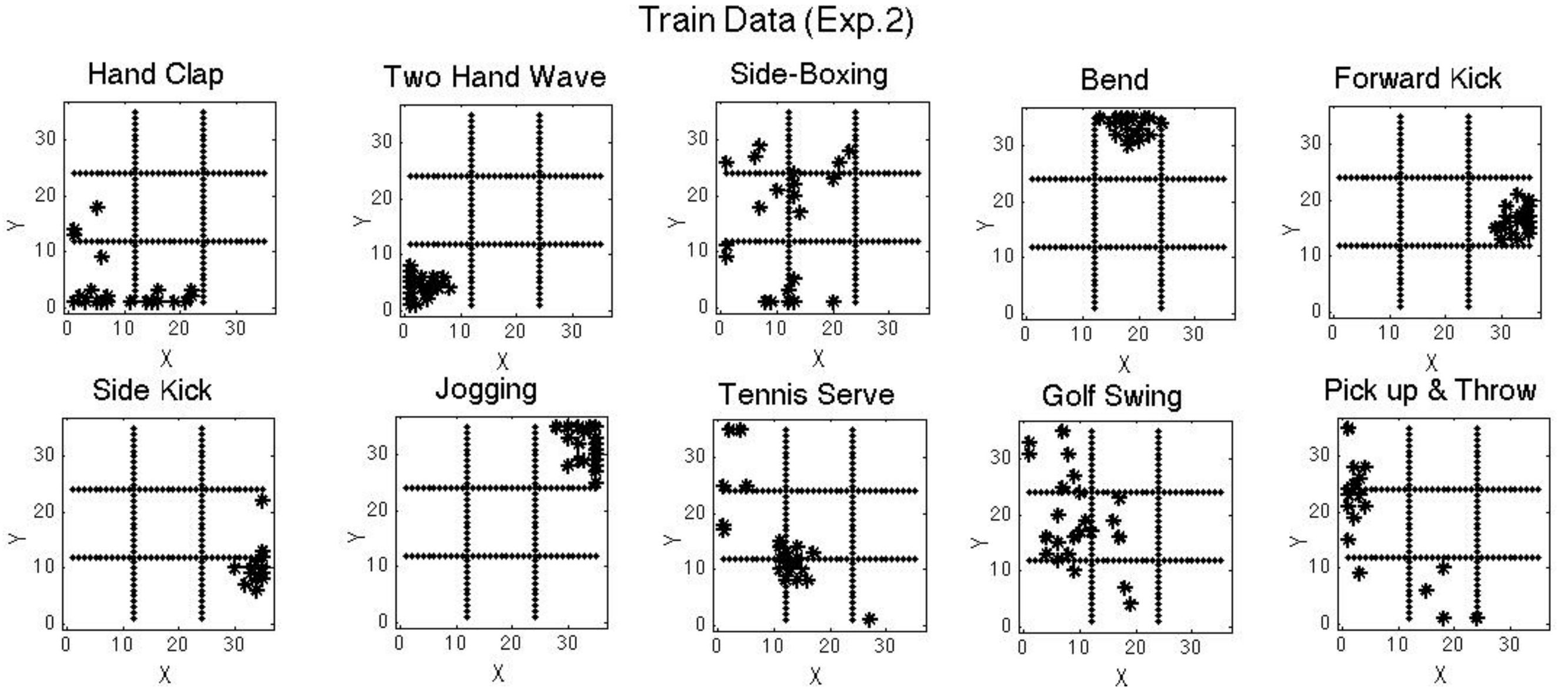}\hspace{5pt}
\caption
{Activations by the training set in the trained second layer SOM in the second experiment. The map is divided into 9 regions, starts from the bottom left square (region 1) and ends in the upper right square (region 9). As can be seen, the activations are spread out over several regions for some actions, but not for others (such as Two Hand Wave, Bend, Forward Kick and Jogging) which are contained in single regions.}
\label{fig:trSOMB}
\end{minipage}
\end{center}
\end{figure}

\begin{figure}%
\begin{center}
\begin{minipage}{120mm}
\centering%
\includegraphics[width=4.70in]{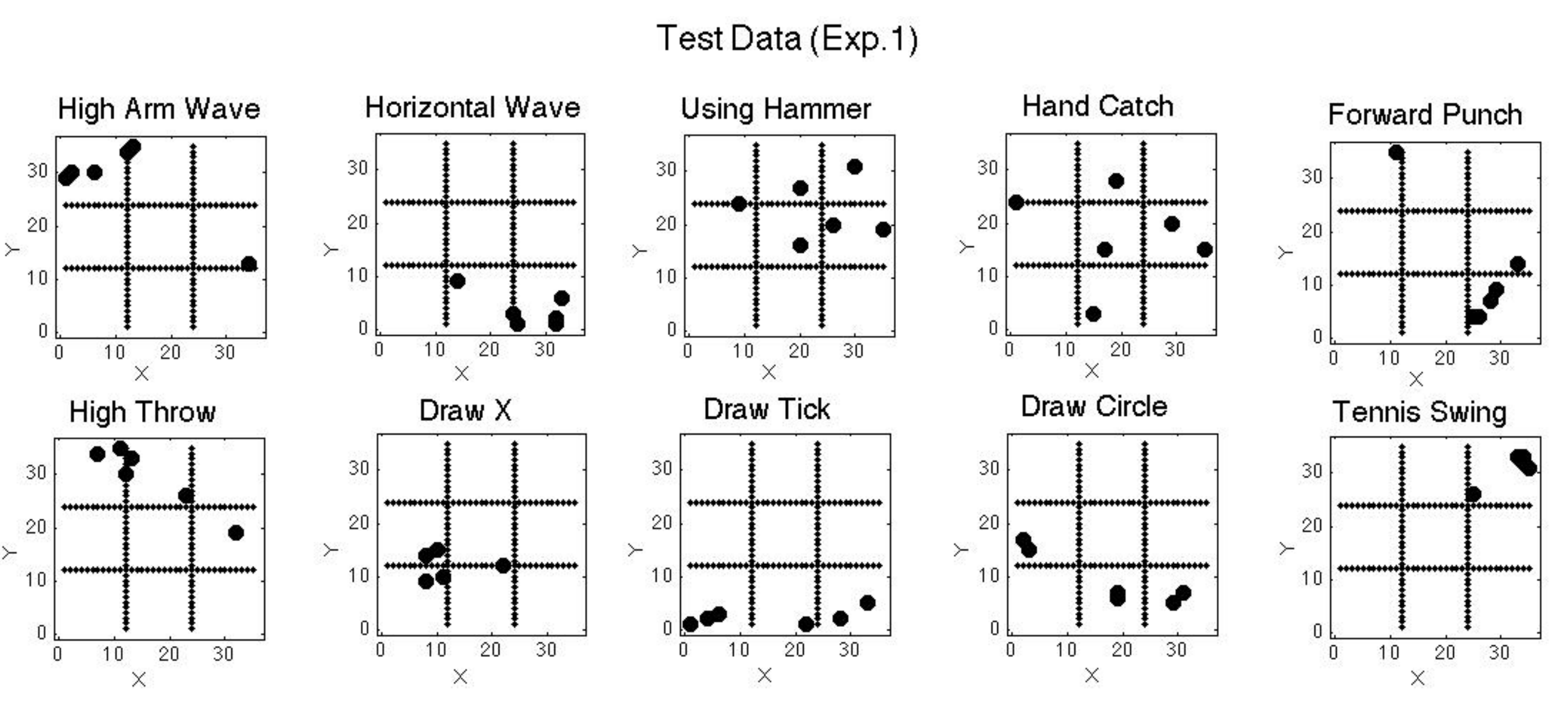}\hspace{5pt}
\caption
{Activations by the test set in the trained second layer SOM in the first experiment. The map is divided into 9 regions, starts from the bottom left square (region 1) and ends in the upper right square (region 9). As can be seen, the activations are spread out over several regions for many actions.}
\label{fig:tsSOMBMap1}
\end{minipage}
\end{center}
\end{figure}

\begin{figure}%
\begin{center}
\begin{minipage}{120mm}
\centering%
\includegraphics[width=4.70in]{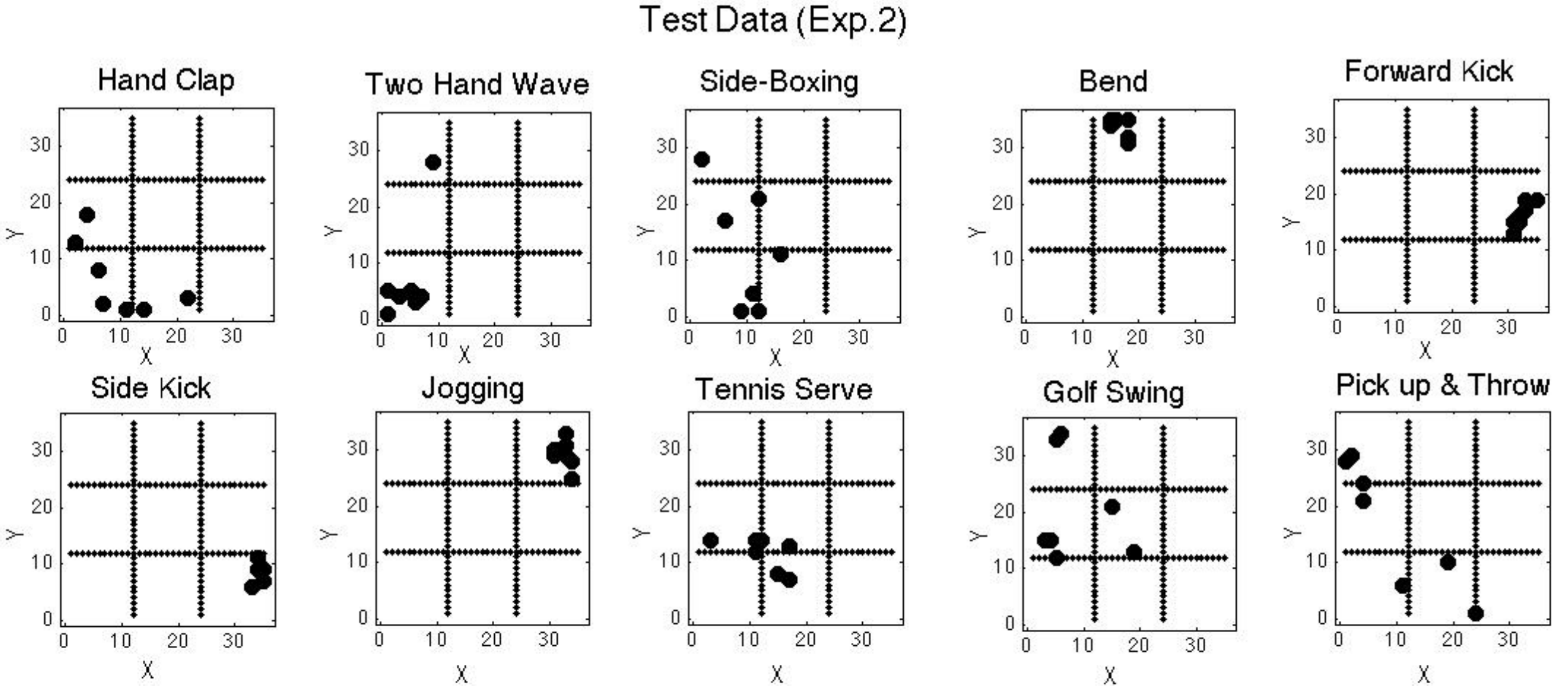}\hspace{5pt}
\caption
{Activations by the test set in the trained second layer SOM in the second experiment. The map is divided into 9 regions, starts from the bottom left square (region 1) and ends in the upper right square (region 9). As can be seen, the activations are spread out over several regions for some actions, but not for others (such as Two Hand Wave, Bend, Forward Kick and Jogging) which are contained in single regions. This is similar as for the training set in the second experiment.}
\label{fig:tsSOMB}
\end{minipage}
\end{center}
\end{figure}

The classification results of the two experiments can be compared by looking at the trained second layer SOMs shown in ~Fig.~\ref{fig:trSOMBMap1} (corresponding to the merged system of the first experiment) and \ref{fig:trSOMB} (corresponding to the merged system of the second experiment). As shown in these figures, for many of the actions there are wide activated areas of neurons. This is especially the situation in the first experiment (see \ref{fig:trSOMB}). This reflects the fact that the actions are performed in multiple styles. This is why an action can be represented in multiple regions in the second layer SOM. This effect can make the classification more difficult due to overlapping of the activated areas of different actions. This makes it important to use a sufficient number of samples of each action performance style to train the system properly and to improve the accuracy. Thus many actions form several sub-clusters in the second layer SOMs, which is depicted in Fig.~\ref{fig:trSOMBMap1} and Fig.~\ref{fig:trSOMB} for the trained data and in Fig.~\ref{fig:tsSOMBMap1} and Fig.~\ref{fig:tsSOMB} for the test data of the first and the second experiment respectively. In these figures we see the neurons activated by each performed action sample. It can be observed that for several actions there are activated neurons in more than one region. 

\begin{figure}%
\begin{center}
\begin{minipage}{120mm}
\centering%
\includegraphics[width=4.50in]{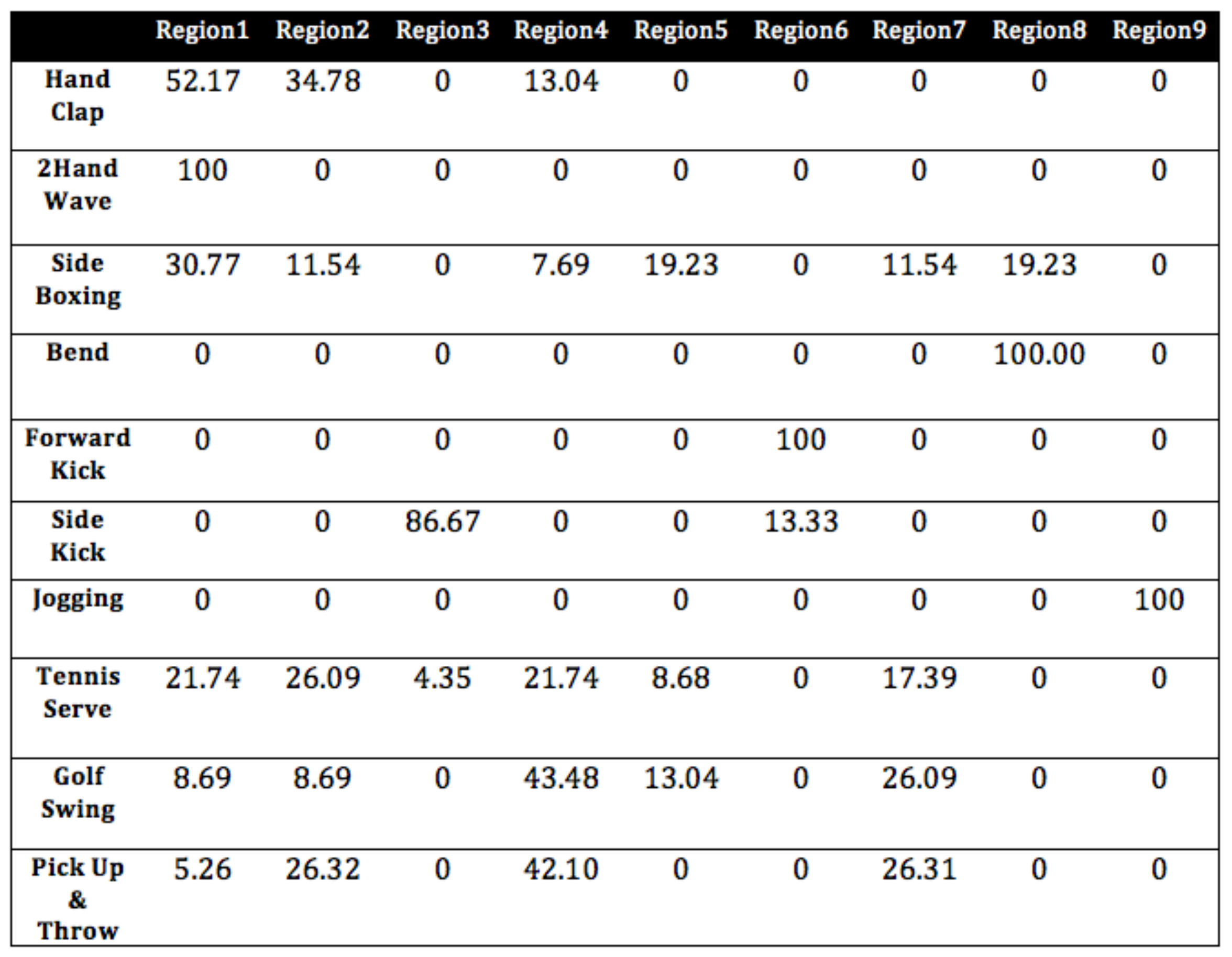}\hspace{5pt}
\caption
{The percentage of activations in each region in the second layer SOM for each action. The values of this table are calculated only for the training data set used in the second experiment as a sample to indicate that the better accuracy in the second experiment might be due to how the second layer SOM is formed.}
\label{fig:table2}
\end{minipage}
\end{center}
\end{figure}

To understand this better, the percentage of activations by an action in each region has been calculated, Fig.~\ref{fig:table2}, for the trained data of the second experiment. By comparing the percentage of activated areas belonging to each action we can see that the actions named Two Hand wave, Bend, Forward Kick and Jogging activated neurons belonging to only one of the 9 regions. This means that these actions can be considered to form only one cluster. For the other actions the representations are spread out in several regions. 

The performance accuracy we have obtained in these experiments can be compared to other relevant studies on the action recognition. In the literature one finds several action recognition systems which are validated and tested on the MSR Action 3D dataset (\cite{MSR}). Our results show a significant accuracy improvement from 74.7\% of the state of the art system introduced in the \cite{WanqingLi}. Even though there is a difference in the way the data set is divided, we can show that our hierarchical SOM architecture outperforms many of the systems tested on the MSR Action 3D data set (such as the systems introduced in \cite{Oreifej}, \cite{Wang3}, \cite{Vieira}, \cite{Wang2}, \cite{Xia1} and \cite{Xia2}).

\begin{figure}%
\begin{center}
\begin{minipage}{120mm}
\centering%
\includegraphics[width=4.50in]{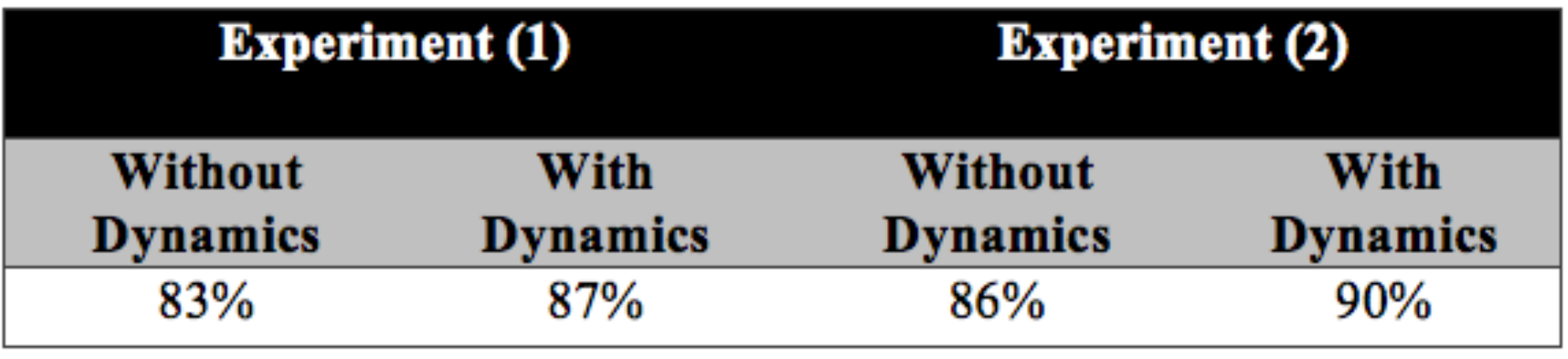}\hspace{5pt}
\caption
{The results of the experiments using MSR Action 3D data with the Hierarchical SOM architecture divided into the results when using postures only and when using postures together with the dynamics.}
\label{fig:Table}
\end{minipage}
\end{center}
\end{figure}

Among the systems using self organizing maps for the action recognition we can refer to \cite{huang} that is a different SOM based system for human action recognition in which a different data set of 2D contours is used as the input data. Although our system and our applied data set are totally different from the ones used in \cite{huang}, our hierarchical SOM architecture outperforms this system too. In \cite{Buonamente3}, a three layer hierarchical SOM architecture is also used for the task of action recognition which has some similarities with the hierarchical SOM architecture presented in this study besides differences in the units such as the pre-processing and ordered vector representation. The system introduced in \cite{Buonamente3} was evaluated on a different dataset of 2D contours of actions (obtained from the INRIA 4D repository) in which the system was trained on the actions performed by one actor (Andreas) and then tested on the actions of a different actor (Helena) which resulted in a performance accuracy of 53\%. We achieved a significant improvement of this result in our hierarchical SOM architecture, which is depicted in the  Fig.~\ref{fig:Table}.  

In our work on action recognition we have also implemented a version of the hierarchical action recognition architecture that works in real time receiving input online from a Kinect sensor with very good results. This system is presented in \cite{Gharaee2}. We have also made another experiment in \cite{Gharaee4} on the recognition of actions including objects in which the architecture is extended to perform the object detection process too. In our future experiments we plan to present our suggested solution for the segmentation of the actions.

\section{Conclusion}
In this article we have presented a system for action recognition based on Self-Organizing Maps (SOMs). The architecture of the system is inspired by findings concerning human action perception, in particular those of \citep {johansson} and a model of action categories from \citet{Gardenfors2}. The first and second layers in the architecture consist of SOMs. The third layer is a custom made supervised neural network.

We evaluated the ability of the architecture to categorize actions in the experiments based on input sequences of 3D joint positions obtained by a depth-camera similar to a Kinect sensor. Before entering the first-layer SOM, the input went through a preprocessing stage with scaling and coordinate transformation into an ego-centric framework, as well as an attention process which reduces the input to only contain the most moving joints. In addition, the first and second order dynamics were calculated and used as additional input to the original joint positions.

The primary goal of the architecture is to categorize human actions by extracting the available information in the kinematics of performed actions. As in prototype theory, the categorization in our system is based on similarities of actions, and similarity is modelled in terms of distances in SOMs. In this sense, our categorization model can be seen as an implementation of the conceptual space model of actions presented in \citet{Gardenfors1} and \citet{Gardenfors2}.

Although categorization based on the first and second order dynamics has turned out to be slightly worse than when sequences of 3D joint positions are used, we believe that this derives from limited quality of the dataset. We have also noticed that the correctly categorized actions in these three different cases do not completely overlap. This has been successfully exploited in the experiment presented in this article, by combining them all to achieve a better performance of the architecture.

Another reason for focusing on the first order dynamics (implemented as the difference between subsequent frames) is that it is a way of modelling a part of the human attention mechanism. By focusing on the largest changes of position between two frames, that is, the highest velocity, the human tendency to attend to movement is captured. We believe that attention plays an important role in selecting which information is most relevant in the process of action categorization, and our experiment is a way of testing this hypothesis. The hypothesis should, however, be tested with further data sets in order to be better evaluated. In the future, we intend to perform such test with datasets of higher quality that also contains new types of actions.

An important aspect of the architecture proposed in this article is its generalizability. A model of action categorization based on patterns of forces is presented in \citep {Gardenfors1} and \citep {Gardenfors2}. The extended architecture presented in this article takes into account forces by considering the second order dynamics (corresponding to sequences of joint accelerations), and, as has been shown, improves the performance. We also think that it is likely that the second order dynamics contains information that could be used to implement automatized action segmentation. We will explore this in the future. The data we have tested come from human actions. The generality of the architecture allows it to be applied to other forms of motion involving animals and artefacts. This is another area for future work.


\bibliography{reference}

\end{document}